# AI Autonomy: Self-Initiated Open-World Continual Learning and Adaptation

Bing Liu,[1] Sahisnu Mazumder,[2] Eric Robertson,[3] Scott Grigsby[3]
[1] University of Illinois at Chicago, USA
[2] Intel Labs, USA
[3] PAR Government Systems Corporation, USA
liub@uic.edu, sahisnumazumder@gmail.com, {eric_robertson, scott_grigsby}@partech.com

## Abstract

As more and more AI agents are used in practice, it is time to think about how to make these agents fully autonomous so that they can (1) learn by themselves continually in a self-motivated and self-initiated manner rather than being retrained offline periodically on the initiation of human engineers and (2) accommodate or adapt to unexpected or novel circumstances. As the real-world is an open environment that is full of unknowns or novelties, the capabilities of detecting novelties, characterizing them, accommodating/adapting to them, gathering ground-truth training data and incrementally learning the unknowns/novelties become critical in making the AI agent more and more knowledgeable, powerful and self-sustainable over time. The key challenge here is how to automate the process so that it is carried out continually on the agent's own initiative and through its own interactions with humans, other agents and the environment just like human on-the-job learning. This paper proposes a framework (called SOLA) for this learning paradigm to promote the research of building autonomous and continual learning enabled AI agents. To show feasibility, an implemented agent is also described.

## 1. Introduction

Classical machine learning (ML) makes the closed-world assumption, which means that the classes of objects seen by the system in testing or application must have been seen during training (Fei, Wang, and Liu 2016; Bendale and Boult 2015; Liu 2020), i.e., there is nothing unexpected or novel occurring in testing or deployment. This assumption is invalid in practice as the real world is an ***open environment*** that is full of unknowns or novel objects. For humans, novelties serve as an intrinsic motivation for learning. Human novelty detection results in a cascade of unique neural responses and behavioral changes that enable exploration and flexible memory encoding of the novel information. As learning occurs, this novelty response is soon lost as repeated exposure to novelty results in fast neural adaptation (Tulving and Kroll 1995; Murty et al. 2013). To make an AI agent thrive in the real open world, like humans, it has to detect novelties and learn them incrementally to make the system more knowledgeable and adaptable over time. It must do so on its own initiative on the job (after deployment) rather than relying on human engineers to retrain the system offline periodically. That is, it must learn in the open world in a self-motivated manner in the context of its performance task (the main task of the agent).

We use the hotel guest-greeting bot example from (Chen and Liu 2018) to illustrate the issues involved. The bot's performance task is greeting hotel guests. When its vision system sees a guest (say, John) it has learned before, it greets him by saying,

"*Hi John, how are you today*?"

When it sees a new guest, it should detect this guest as new or novel. This is a ***novelty detection*** problem (also known as ***out-of-distribution*** **(OOD) detection**). upon discovering the novelty - the new guest, it needs to ***accommodate*** or ***adapt*** to the novel situation. The bot may say to the new guest,

"*Hello, welcome to our hotel! What is your name, sir*?"

If the guest replies "David," the bot takes some pictures of the guest to ***gather training data*** and then ***incrementally or continually learn*** to recognize David. The name "David" serves as the class label of the pictures taken. As for humans, the detected novelty serves as an intrinsic self-motivation for the agent/bot to learn. When the bot sees this guest again next time, it can say

"*Hi David, how are you today*?" (David is no longer new)

In an actual hotel, the situation is, however, much more complex than this. For example, how does the system know that the novel object is a person, not a dog? If the system can recognize the object as a person, how does it know that he/she is a hotel guest, not a service provider for services such as delivery or security? To adapt to the novel object or situation, the system must first ***characterize*** the novel object, as without it, the agent wouldn't know how to ***adapt*** or ***respond***. In this case, some classification or similarity comparison is needed to decide whether it is a person with luggage. If the object looks like a person but has no luggage, the bot will not respond or learn to recognize the person as it is ***irrelevant*** to its performance task. If the novel object looks like an animal, it should notify a hotel employee and learn to recognize the object so that it will no longer be novel when it is seen next time. In short, for each characterization, there is a corresponding response or adaptation strategy, which can be NIL (i.e., do nothing). This discussion shows that to characterize, the agent must already have rich world knowledge. Finally, there is also ***risk*** involved when making an incorrect decision.

As classic learning matures, we should go beyond the existing paradigm to study how to enable an agent to learn and adapt by itself via its own interactions with humans and the environment, i.e., self-initiation, involving no engineers. This paper proposes a ***S****elf-initiated* ***O****pen-world continual* ***L****earning and* ***A****daptation* (**SOLA**) framework to promote the research of autonomous AI agents so that they can face the real open world and learn by themselves. An example SOLA agent in the context of dialogue systems or chatbots that implements the SOLA framework is also discussed.

## 2. Comparison with Related Work

Open world learning has been studied by many researchers (Bendale and Boult 2015; Fei, Wang, and Liu 2016; Xu et al. 2019), but they mainly focused on novelty detection (Parmar et al. 2021), also called *open set* or *out-of-distribution* (OOD) detection. Some researchers have also studied learning the novel objects after they are detected (Bendale and Boult 2015; Fei, Wang, and Liu 2016; Xu et al. 2019) and manually labeled. A survey of the topic can be found in (Yang et al. 2021). A position paper (Langley 2020) recently presented some blue-sky ideas about open world learning, but it does not have sufficient details or an implemented system. SOLA differs from these prior studies in many ways,

**(1)** SOLA stresses "self-initiation" in learning, which means that all the learning activities from start to end are self-motivated and self-initiated by the agent itself. The process involves no human engineers.

**(2)** Due to self-initiation, SOLA enables learning after model deployment like human learning on the job or while working, which has barely been attempted before. In existing learning paradigms, after a model has been deployed, there is no more learning until the model is updated or retrained on the initiation of the human engineers.

**(3)** SOLA is a lifelong and continual learning paradigm again because learning is self-initiated and unceasing. It is thus connected with lifelong and continual learning, which is an active research area in machine learning, computer vision and natural language processing (Chen and Liu 2018).

**(4)** SOLA involves online interactions of the learning agent with human users, other AI agents, and the environment. The purpose is to acquire ground-truth training data on the fly by itself (and it is free). This is very similar to what we humans do when we encounter something novel or new and ask others interactively to acquire knowledge. It is very different from collecting a large amount of unlabeled data and asking human annotators to label the data (as in the case of crowdsourcing). Also, it differs from active learning (Settles 2009; Ren et al. 2021) as active learning only focuses on acquiring labels from users for selected unlabeled examples in the given dataset. SOLA also allows learning from other resources, e.g., the Web or a human teacher, to gain knowledge, like a human reading a book or learning from a dedicated teacher. Due to space limits, this paper will not focus on these types of learning [see an example in (Mitchell et al. 2015; Kasaei et al. 2020)].

(5) SOLA includes modules to characterize and to adapt to novel situations so that the agent can work in the open world environment and also continually learn and self-sustain in the process.

In robotics research, a closely related topic is open-ended learning, where the number of classes or categories to be learned is not predefined. In (Seabra Lopes, and Chauhan 2007, 2008), a human teacher interactively teaches a robot to learn new words or new object names through a user-interface. In the process, the robot learns incrementally. In (Kasaei et al. 2020; Kasaei et al. 2019), a user teaches a robot to incrementally learn to recognize prior unknown visual objects and their affordance categories. The system in (Oliveira et al. 2015) also learns attributes or codebook words used to encode the objects. In all these cases, the *teacher/user takes the initiative and decides what to teach and how to teach*. In the proposed SOLA framework, we focus on the system itself. The system takes the initiative to detect what it does not know or what is novel and to incrementally learn the new/novel objects through interaction with humans or the environment. The SOLA framework also includes the characterization of the new objects so that the system can formulate a plan of actions to respond to the novel/new objects. In the process, the agent also considers risk and safety. The SOLA framework thus covers open-ended learning as it also has interactive learning and continual/incremental learning.

Intrinsically motivated open-ended learning (IMOL) is also related. It aims to develop robots that can autonomously generate internal motivational signals or rewards to acquire knowledge and skills (Barto et al., 2004; Oudeyer et al., 2007; Santucci 2020; Mirolli and Baldassarre, 2013). It was inspired by the ability of humans to discover interesting things to learn driven by self-generated rewards or curiosity not related to any specific external tasks (White, 1959; Deci and Ryan, 1985). Knowledge-based intrinsic motivations (IMs) (Oudeyer and Kaplan 2007) are more closely related to our work, which are of two types, namely *novelty-based IMs* and *prediction-based IMs* (Barto et al. 2013; Baldassarre 2019). Novelty-based IMs try to detect novel items and direct attention to novel items in exploration. Prediction-based IMs try to predict the future and compare with the observed reality to compute the prediction error, and then direct the attention to the wrongly predicted items to improve the prediction accuracy. IMs are often expressed as internal rewards to augment sparse external rewards in reinforcement learning (Kulkarni et al., 2016; Pathak et al. 2017; Baldassarre 2019). Although related, IMOL is significantly different from SOLA. IMOL has the ambitious goal of imitating humans' cognitive and learning process and capability in robots, while SOLA's goal is more modest in the sense that it has well-defined novelty detection, continual learning and other computational functions. SOLA is not normally associated with reinforcement learning, and novelty in SOLA serves as the motivation for continual learning and triggers self-initiation.

Human-robot teaming also has some resemblance to our work. For example, Talamadupula et al. (2017) proposed a system that enables human-robot interactions through natural language dialogues to jointly perform a task. The term "open world" in this paper means that there may be new goals, new sub-tasks, and new entities in the task, e.g., a search and rescue task. Apart from natural language dialogue, the system also performs some reasoning and open-world planning. However, this work does not involve continual learning or knowledge accumulation, which is the core of the proposed SOLA framework.

In summary, SOLA makes learning autonomous and self-initiated. Although novelty detection, adaptation and continual learning have been studied discretely (in specific use cases or application scenarios) in many works over decades, we have not found any work that has discussed or provided a generic and holistic framework (like SOLA) that unifies the ideas of self-initiation, novelty detection, adaptation, and open-world continual learning into one. We believe that SOLA is necessary for the next generation machine learning and AI agents. Finally, note that although SOLA focuses on self-initiated learning, it does not mean that the learning system cannot learn a task given by humans or other AI agents (as followed in usual ML design practices).

## 4 Novelty Detection

Novelty is a core concept of SOLA as it triggers and motivates the whole SOLA process. Detecting novel objects or situations is thus a critical task. In the research community, it is often called *out-of-distribution* (OOD) *detection*. In general, *novelty* is an agent-specific concept. An object may be novel to one agent based on its partial knowledge of the world but not novel to another agent. We distinguish two types of novelty, *absolute novelty* and *contextual novelty*.

**Absolute novelty.** Absolute novelty represents something that the agent has never seen before. For example, in the context of supervised learning, the agent's world knowledge is learned from the training data $D_{tr} = \{(x_i, y_i)\}_{i=1}^{n}$ with $x_i \in X$ is the input data and $y_i \in Y_{tr}$ is its class label. Let $h(x)$ be the latent or internal representation of $x$ in the agent's mind, $h(D_{tr}^{i})$ be the latent representation of the training data of class $y_i$, and $k$ $(= |Y_{tr}|)$ be the total number of training classes. We use $\mu(h(x'), h(D_{tr}^{i}))$ to denote the novelty score of a test instance $x'$ with respect to $h(D_{tr}^{i})$. The degree of novelty of $x'$ with respect to $D_{tr}$, $\mu(h(x'), h(D_{tr}))$, is defined as the minimum novelty score with regard to every class,

$$\mu(h(x'), h(D_{tr})) = \min\{\mu(h(x'), h(D_{tr}^{1})), \ldots, \mu(h(x'), h(D_{tr}^{k}))\} \quad (1)$$

The novelty function $\mu$ can be defined based on specific applications. For example, if the training data of each class follows the Gaussian distribution, one may use z-score or Mahalanobis distance as the novelty score. The definition is also for ***out-of-distribution*** (**OOD**) *detection*, where the training classes in $Y_{tr}$ are called ***in-distribution*** **(IND)** classes and those test instances that do not belong to the IND classes are called OOD instances. Thus, an OOD or novelty detection model can classify a test instance from an IND class to its corresponding class and detect OOD test instances that do not belong to any IND class.

*Novel instance*: A test instance $x'$ is novel if its novelty score $\mu(h(x'), h(D_{tr}))$ is greater than or equal to a threshold value γ such that $x'$ can be assigned a new class that is not in $Y_{tr}$.

*Novel class*: A newly created class $y_{new}$ ($y_{new} \notin Y_{tr}$) assigned to some novel instances is called a *novel class* (unknown or unseen class). The classes in $Y_{tr}$ are also called known or *seen classes*.

**Contextual novelty.** Given that both the instance x and the context *Q* are not absolutely novel and the probability *P*(*x*|*Q*) of x occurring in *Q* is very low, but *x* has occurred in *Q*, which is *surprising* or *unexpected*. A contextual novelty is also commonly called a ***surprise*** or ***unexpected event***. In human cognition, surprise is an emotional response to an instance that greatly exceeds the expected uncertainty within the context of a task. The definitions of contextual novel instance and class are like those for absolute novelty.

Intuitively, in absolute novelty, the novelty of *x* is context independent. For example, if the agent has never seen a tiger before, it is absolutely novelty, irrespective of its context (other objects in the image). If *x* is known but is very unlikely to appear in a context, *x* is contextually novel, e.g., a deer (known) appears in a crowded city street. A related work on contextual novelty detection can be found in (Ma et al., 2021; 2022).

Novelty is not restricted to the perceivable physical world but also includes the agent's internal world, e.g., novel interpretations of world states or internal cognitive states that have no correspondence to any physical world state. Interested readers may also read (Boult et al. 2021) for a more nuanced and perception-based study of novelty. There are other related concepts to novelty, e.g., *out-of-distribution* (OOD) *samples*, *outliers,* and *anomalies*. An extensive work has been done on novelty detection (Yang et al. 2021).

**Outlier and anomaly:** An outlier is a data point that is far away from the main data clusters, but it may not be unknown. For example, the salary of a company CEO is an outlier with regard to the salary distribution of the company employees, but that is known and thus not novel. Unknown outliers are novel. Anomalies can be considered outliers or instances that are one off and never repeated. Though technically "novel" they may not result in a new class. Note that this paper does not deal with various types of data shift such as covariate shift, prior probability shift and concept drift as a large amount of work has been done (Moreno-Torres et al. 2012). We will not discuss novelty detection further because it has been studied extensively (Pang et al. 2021; Parmar et al. 2021; Yang et al. 2021).

# 4. Lifelong and Continual Learning

Since SOLA at its core is a continual learning paradigm, this section introduces lifelong or continual learning (Chen and Liu 2018). To enable autonomous continual learning without the involvement of human engineers, other capabilities are needed, which we will discuss in subsequent sections. The terms lifelong learning and continual learning have the same meaning and are used interchangeably.

## 4.1 Continual Learning (CL)

CL is defined in (Chen and Liu 2018) as follows, which is based on the early definitions in (Thrun 1995; Silver, Yang, and Li 2013; Ruvolo and Eaton 2013; Chen and Liu. 2014; Mitchell et al. 2015) and more recent research (Rusu et al. 2016; Kirkpatrick et al. 2017; Zenke, Poole, and Ganguli 2017; Rebuffi, Kolesnikov, and Lampert 2017; Shin et al. 2017; Serra et al. 2018; Lee, Stokes, and Eaton 2019; Chaudhry et al. 2020; Ke, Liu, and Huang 2020; Ke et al. 2021; Guo et al. 2022; Kim et al. 2022b).

**Definition:** *Continual learning* (CL) aims to learn a sequence of tasks. At any point in time, the learner is assumed to have $N$ tasks, $T_1, T_2, \ldots, T_N$ (called the previous tasks). In learning the $(N + 1)^{th}$ task $T_{N+1}$ (called the new task or the current task), the learner wants to achieve two main objectives:

(1) *Overcoming catastrophic forgetting* (CF). CF refers to the phenomenon that when a neural network learns a sequence of tasks, the learning of each new task is likely to change the weights learned for previous tasks, which degrades the model performance for the previous tasks (McCloskey and Cohen 1989).

(2) *Encouraging knowledge transfer* (KT) *across tasks*. The learner should leverage the knowledge in the knowledge base (KB) to help learn $T_{N+1}$. This is called the *forward transfer*. The new task should also help improve some previous task models if possible. This is call *backward transfer*. An explicit or implicit knowledge base (KB) is maintained to retain the knowledge learned from the previous $N$ tasks. After the completion of learning $T_{N+1}$, KB is updated with the knowledge gained from $T_{N+1}$.

Two CL settings have been studied extensively in the research literature. See (Kim et al. 2022b) for their formal definitions.

**Class incremental learning (CIL).** In CIL, each task consists of one or more classes to be learned together but only one model is learned to classify all classes learned so far. In testing, a test instance from any class may be presented to the model for it to classify with no task related information provided.

**Task incremental learning (TIL).** In TIL, each task is a separate classification problem (e.g., one classifying different breeds of dogs and one classifying different types of animals). TIL builds a set of classification models (one per task) in a shared neural network. In testing, the system knows to which task each test instance belongs and uses only the model for the task to classify the test instance.

Earlier research mainly focused on KT in TIL and assumed that the tasks are similar, which clearly facilitates KT across tasks (Chen and Liu 2018). Little work was done on CF, which has been researched only after deep learning became popular. More recent research focused on both CF and KT. When the tasks are similar, KT is the focus (Ke et al. 2021). When the tasks are dissimilar overcoming CF is the key (Chen and Liu 2018; Guo et al. 2022). Work has also been done to learn a mixed sequence of similar and dissimilar tasks, which must deal with both CF and KT at the same time (Ke, Liu, and Huang 2020), i.e., to perform selective knowledge transfer among similar tasks and also to overcome CF for dissimilar tasks. Task similarity is detected automatically.

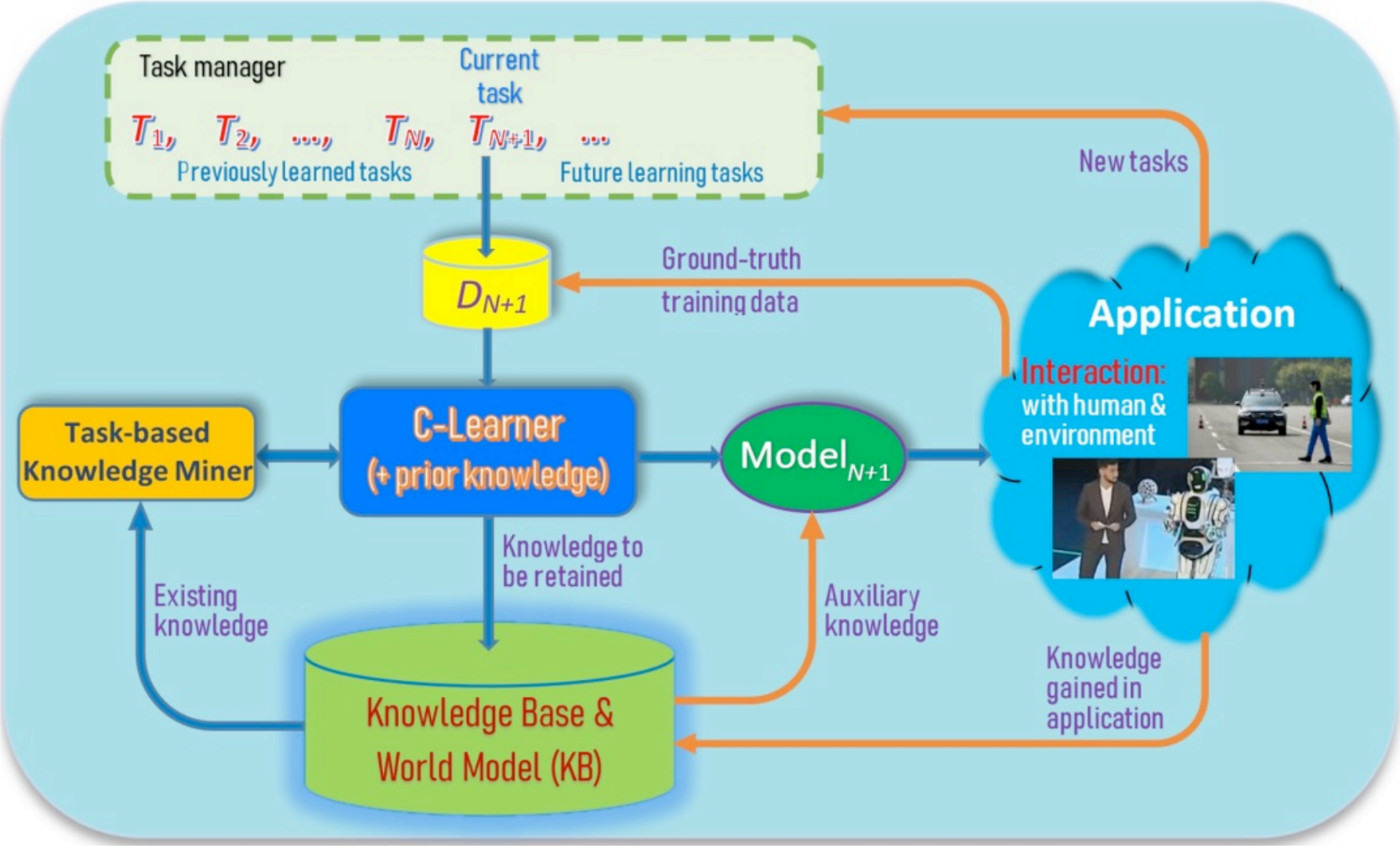

Figure 1: Architecture of a typical continual learning framework incorporating on-the-job learning (best viewed in color). $T_1$, ..., $T_N$ are the previously learned tasks, $T_{N+1}$ is the current new task to be learned and DN+1 is its training data. The C-Learner (continual learner) learns by leveraging the relevant prior knowledge identified by the Task-based Knowledge Miner from the Knowledge Base (KB), which contains the knowledge retained in the past. It also deals with CF. Existing research on continual learning does not have the orange-colored lines. The orange-colored lines are added for on-the-job learning in SOLA.

The architecture of CL systems is given in Figure 1 without the orange-colored links. The orange-colored links will be discussed in the next subsection. Dealing with CF is not reflected in the architecture as it stays in the continual learner (C-Learner). $Model_{N+1}$ includes all the models from $T_1$ to $T_{N+1}$, which may all be in one neural network. In the case of TIL, they may be separate models indexed by their task identifiers. In the case of CIL, $Model_{N+1}$ is a single model that covers all classes of the tasks learned so far. It has been proven in (Kim et al. 2022b) that strong *out-of-distribution* (OOD) *detection* or *novelty detection* is a necessary condition for achieving good CIL performances. Apart from overcoming CF and encouraging KT in CIL, this paper also identified a new challenge of *inter-task class separation* (ICS) that is as hard to deal with as CF and KT (see (Kim et al 2022b to details)).

**Limitation.** One key limitation of the existing CL paradigm is that the tasks and their training data are given by the user or engineers. This means that the system is not autonomous and cannot learn by itself. In order to do that, we extend the CL architecture with the orange-colored links in Figure 1 to enable *learning on the job* to achieve the full SOLA.

## 4.2. A New CL Architecture

The new architecture is summarized in the full diagram in Figure 1 with the orange-colored links added to the traditional CL. These links enable the system to learn by itself to achieve autonomy in the SOLA framework, i.e., to learn on the job during application or after model deployment. However, as discussed in the Introduction section, learning during application is more complicated and the framework also includes other aspects related to *adaptation* to the novelty, which we will detail in Section 5 with a new figure showing the full architecture of SOLA.

The basic idea for learning in SOLA is that during application, if the system/agent encounters anything that is out-of-distribution or novel (a novelty), the system first creates a new task to learn and then obtains the needed ground-truth training data to learn the task on the initiation of the system itself through interactions with the humans and the environment. Some new knowledge (or auxiliary knowledge) gained from the application can be added to the KB that may be leveraged in future learning or to improve the current model.

# 5 The Proposed SOLA Framework

The SOLA architecture is given in Figure 2, which adds and elaborates the orange-colored links and associated components to the traditional CL architecture (without the orange-colored links in Figure 1). These newly added links and components enable the system to learn by itself and to adapt to the new situation to gain autonomy, which is what SOLA aims to achieve. It is called **learning after deployment** or **learning on the job** during application or after model deployment.

**Learning after deployment** refers to learning continuously after the model has been deployed in an application or during model application (Liu and Mazumder 2021) on the fly. The basic idea is that during application, if the system/agent encounters anything that is out-of-distribution (OOD) or novel, it needs to **detect the novelty**. Based on the novelty, the system **creates a new task** to learn and acquires the **ground-truth training data** to learn the task on the initiation of the system itself through interactions with humans, other agents and/or the environment. The system then **learns the new task incrementally** or continually. In the process, the system also **adapts itself to the new situation** and carries on its performance task. The whole process is carried out on the fly during application.

## 5.1. Components of SOLA

SOLA is proposed as a framework for building autonomous AI agents. An AI agent consists of a pair of key modules ($P$, $S$), where $P$ is the primary *task-performer* that performs its performance task (e.g., the dialogue system of the greeting bot) and $S$ is a set of *supporting or peripheral functions* (e.g., the vision system and the speech system of the bot) that supports the primary task-performer. The primary task-performer $P$ or each supporting function $S_i \in S$ consist of eight core sub-systems ($L$, $K$, $M$, $R$, $C$, $A$, $S$, $I$). Figure 2 shows the relationships and functions of the sub-systems. We do not distinguish $P$ and $S_i$ in terms of techniques or subsystems as we believe they have no fundamental difference.

- $L$ is an **OWC-Learner** (*Open-World Continual Learner*) that builds models to not only classify the input into known classes but also detect novel objects that have not been seen in training. For example, for the greeting bot, $L$ of the primary task performer $P$ which is a continual learning dialogue system similar to that in Section 6. For the supporting vision system, $L$ can learn to recognize guests and detect novel or unknown objects. Compared to C-Learner in continual learning in Figure 1, OWC-Learner in SOLA not only can learn continually like C-Learner but also produce models that can detect novel instances in testing or in application deployment (see (Kim et al. 2022a) for an example system).
- $K$ is the **Knowledge Base & World Model** (KB) that is important for the performance task, supporting functions and/or the OWC-Learner. KB is also in Figure 1 but plays more roles in SOLA. Apart from keeping the learned or prior knowledge of the domain and the world model, if needed, reasoning capability may also be provided to help the other modules of the system (see the orange-colored links). Some knowledge from the application observed by the Adaptor (see below) may be added to the KB, which can provide some knowledge to the Model for its decision making. World model refers to the representation of the task environment and the commonsense knowledge about the objects and their relationships within.
- $M$ is the **Model** learned by $L$. $M$ takes the input or perception signals from the application environment and make a decision to perform actions in the application. The decision making involves detecting novel or normal input instances, besides performing its intended task (on normal instances) as defined in the application. *M* may also use some input or knowledge from the other supporting functions and task-specific auxiliary knowledge from *K* to help the inference process.

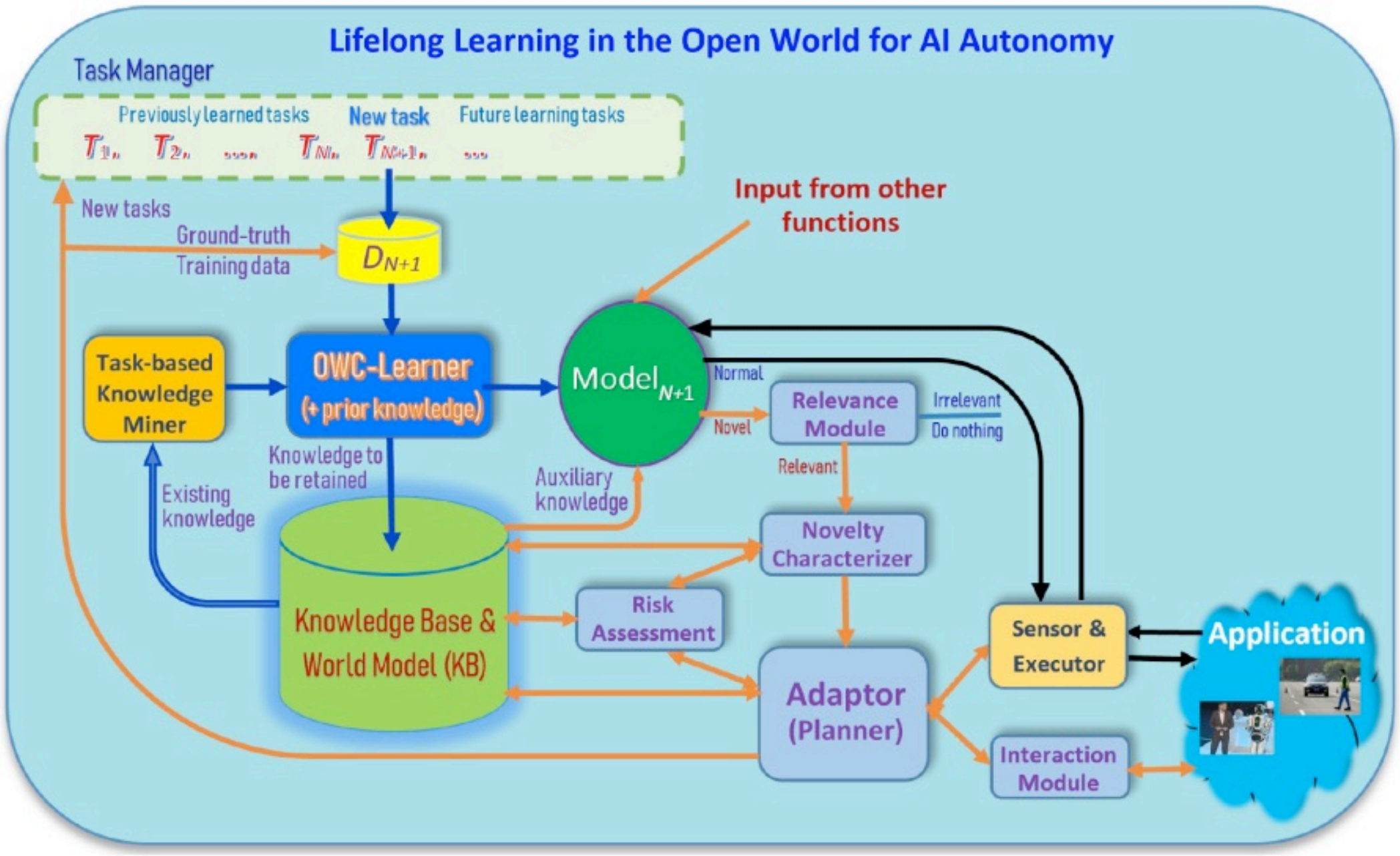

Figure 2: Architecture of the primary task performer or any supporting function. OWC-Learner means Open-World Continual Learner.

- *R* is the **Relevance Module** or focusing mechanism that decides whether the detected novelty is relevant to the current task or not. If it's relevant, the agent should respond to the novelty (discussed below); otherwise simply ignore it. For example, in the greeting bot application, if the bot sees a new inanimate object in its viewing zone, even if the object is novel (never seen before), the bot should ignore it as an inanimate object should not be greeted! So, novelty in inanimate objects is considered as irrelevant.
- *C* is the **Novelty Characterizer** that characterizes the detected novelty based on the knowledge in the KB so that the adaptor (below) can formulate a course of actions to respond or adapt to the novelty. For the characterizer *C* of *P* of the greeting bot, as *P* is a dialogue system, when it cannot understand the utterance of a hotel guest (a novelty), it should decide what it can and cannot understand (see Section 6) and ask the guest based on its partial understanding (see below). In the case of the supporting vision system, when a novel object it detected, the charaterizer may decide what the object looks like and its physical attributes. For example, the novel object may look like a dog based on the greeting bot's KB. (see Section 5.4 for more discussions).
- *A* is the **Adaptor** that adapts to or accommodates the novelty based on the characterization result. It is a ***planner*** that produces a plan of actions for the executor *E* (it can be a task user interface, not considered as a core component of SOLA) or the interactive module *I* to perform. The goal here is to formulate a strategy to respond to the novelty (e.g., acquiring knowledge to learn the detected novelty or reporting it to some other agents in the environment). Given the characterization (e.g., partial understanding) above, *A* may adapt by asking the guest to clarify (see Section 6) and then learn to understand the utterance. In the case of the vision system, if the characterizer believes that the novel object looks like a dog, the adaptor may decide to report to a hotel employee and then learns the new object by taking some pictures as the training data. It can also ask the hotel employee for the name of the object as the class label. In the latter two cases, *A* needs to invoke *I* to interact with the human and *L* to learn the novelty so that it will not be novel in the future. That is, *A* is also responsible for creating new tasks (e.g., learning to recognize new objects by the greeting bot) on the fly and proceeds to acquire ground truth training data with the help of $I$ (discussed below) to be learned by $L$. This adaptation process often involves reasoning and may utilize knowledge from *K*.

- $S$ is the **Risk Assessment** module. Novelty implies uncertainty in adapting to the novel situation. In making each response decision, risk needs to be assessed (see Section 5.5 for more discussions).
- *I* is the **Interactive Module** for the agent to communicate with humans or other agents, e.g., to acquire ground-truth training data or to get instructions when the agent does not know what to do in a unfamiliar situation. It may use the natural language (for interaction with humans) or an agent language (for interaction with other agents) for communication.

Several remarks are in order. **(1)** not all agents need all these sub-systems, and some sub-systems may also be shared. For example, the primary task performer *P* in the greeting bot application is a dialogue system. Its interaction module *I* can use the same dialogue system. In some cases, the model *M* may also be able to determine the relevance of a novel object to the application and even characterize the novelty because characterization in many cases is about classification and similarity comparison. **(2)** as we will see, every sub-system can and should have its own local learning capability. **(3)** the interaction module *I* and the adapter *A* will create new tasks to learn and gather ground truth training data for learning. **(4)** most links in Figure 2 are bidirectional, which means that the sub-systems may need to interact with each other to perform their tasks. The interactions may involve requesting for information, passing messages, and/or going back and forth with hypothesis generation, revision and evaluation to make more informed decisions.

Since the primary task performer *P* and each supporting sub-system $S_i$ has the same components or sub-systems, we will discuss them in general rather than distinguishing them.

## 5.2. Open World Continual Learning

The classical machine learning (ML) makes the i.i.d assumption, which is often violated in practice. Here we first define several related concepts and then the idea of *open world continual learning* in SOLA.

Let the training data that have been seen so far from previous tasks be $D_{tr} = \{(x_i, y_i)\}_{i=1}^{n}$ with $x_i \in X$ as the input data and $y_i \in Y_{tr}$ as its class label. Let the set of class labels that may appear in testing or application be $Y_{tst}$. Classical ML makes the *closed-world assumption*.

**Closed-world assumption:** There are no new or novel instances or classes that may appear in testing or application, i.e., $Y_{tst} \subseteq Y_{tr}$. In other words, every class seen in testing or application must have been seen in training.

**Open world:** There are test classes that have not been seen in training, i.e., $Y_{tst} - Y_{tr} \neq \emptyset$.

**Definition** (**closed-world learning**): It refers to the learning paradigm that makes the closed-world assumption.

**Definition** (**open world learning**): It refers to the learning paradigm that performs the following functions: (1) classify test instances belonging to training classes to their respective classes and detect novel or out-of-distribution instances, and (2) learn the novel classes labeled by humans for the identified novel instances to update the model using the labeled data. The model updating is initiated by human engineers and involves re-training or incremental learning.

**Definition** (**SOLA**): SOLA is a learning paradigm that performs open-world learning but the learning process is initiated by the agent itself after deployment with no involvement of human engineers. The new task creation and ground-truth training data acquisition are done by the agent via its interaction with the user and the environment. The learning of the new task is incremental, i.e., no re-training of previous tasks/classes. The process is lifelong or continuous, which makes the agent more knowledgeable over time. In addition to learning, SOLA also characterizes and adapt/respond to the novelty so that its performance task can still be carried out without stopping.

**Steps in learning in SOLA.** The main continual learning process in SOLA involves the following three steps, which can be part of the novelty adaptation or accommodation (see Section 5.5).

**Step 1** - *Novelty detection*. This step detects data instances whose classes do not belong to $Y_{tr}$. A fair amount of research has been done on this (see the surveys (Pang et al. 2021; Parmar et al. 2021; Yang et al. 2021)).

**Step 2** - *Acquiring class labels and creating a new learning task on the fly*: This step first clusters the detected novel instances. Each cluster represents a new class. It may be done automatically or through interactions with humans using the interaction module *I*. Interacting with human users should produce more accurate clusters and obtain meaningful class labels. If the detected data is insufficient for building an accurate model to recognize the new classes, additional ground-truth data may be collected via interaction with users (and/or passively by downloading data from the Web like searching and scrapping images of objects of a given class). A new learning task is then created.

In the case of our hotel greeting bot, since the bot detects a single new guest (automatically), no clustering is needed. It then asks the guest for his/her name as the class label. It also takes more pictures as the training data. With the labeled ground-truth data, a new learning task is created to incrementally learn to recognize the new guest on the fly.

The learning agent may also interact with the environment to obtain training data. In this case, the agent must have an ***internal evaluation system*** that can assign rewards to different states of the world, e.g., for reinforcement learning.

**Step 3** - *Incrementally learn the new task*. After ground-truth training data has been obtained, the learner *L* incrementally learns the new task. This is continual learning (Chen and Liu 2018). We will not discuss it further as there are already numerous existing techniques (Parisi et al. 2019; Kim et al. 2022b; Lomonaco et al. 2022). Many can leverage existing knowledge to learn the new task better (Chen and Liu 2018).

## 5.3. Relevance of Novelty

Due to the performance task, the agent should focus on novelties that are critical to the performance task. For example, a self-driving car should focus on novel objects or events that are or may potentially appear on the road in front of the car. It should not pay attention to novel objects in the shops along the street (off the road) as they do not affect driving. This relevance check involves gathering information about the novel object to make a classification decision.

## 5.4. Novelty Characterization and Adaptation

In a real-life application, classification may not be the primary task of an agent. For example, in a self-driving car, object classification supports its primary performance task of driving. To drive safely, the car must take actions to adapt or respond to the novel/new objects, e.g., slowing down and avoiding the objects. To know what actions to take to adapt, the agent must characterize the new object. The ***characterization*** of a novel object is a description of the object based on the agent's existing knowledge of the world and/or description of agent's uncertainty about the object. Based on the characterization, appropriate actions are formulated to ***adapt*** or respond to the novel object. The process may also involve learning.

*Novelty characterization and adaptation* (or response) form a pair (*c*, *r*), where *c* is the characterization of the novelty and *r* is the adaptation response to the novelty, which is a plan of dynamically formulated actions based on the characterization of the novelty. The two activities go together. If the system cannot characterize a novelty, it takes a low risk-assessed default response. In our greeting bot example, when it can characterize a novelty as a new guest, its response is to say "*Hello, welcome to our hotel! What is your name, sir*?" If the bot has difficulty with characterization, it can take a default action, e.g., '*do nothing*.' The set of responses are specific to the application. For a self-driving car, the default response to a novel object is to slow down or stop the car so that it will not hit the object.

In some situations, the agent must take an action under low confidence circumstances, the agents engage in reinforcement learning, i.e., trying actions and then assessing outcomes.

Characterization can be done at different levels of detail, which may result in more precise or less precise responses. Based on an ontology and object attributes related to the performance task in the domain, the characterization can be described based on the ***type of the object*** and the ***attribute of the object***. For example, in the greeting bot application, it is useful to determine whether the novel object is a human or an animal because the responses to them are different. For self-driving cars, when sensing a novel object on the road, it should focus on those aspects that are important to driving, i.e., whether it is a still or a moving object. If it is a moving object, the car should determine its direction and speed of moving. Thus, the classification of movement is needed in this case to characterize the novelty, which, in turn, facilitates determination of the agent's responding action(s). For instance, if the novel object is a mobile object, the car may wait for the object to leave the road before driving.

Another characterization strategy is to ***compare the similarity*** between the novel object and the existing known objects. For example, if it is believed that the novel object looks like a dog (assuming the agent can recognize a dog), the agent may react like when it sees a dog on the road.

**Dealing with characterization and adaptation.** The above discussion implies that to effectively characterize a novelty, the agent must already have a great deal of world knowledge that it can use to describe the novelty. Additionally, the characterization and response processes are often interactive in the sense that the agent may choose a course of actions based on the initial characterization. After some actions are taken, it will get some feedback from the environment. Based on the feedback and the agent's additional observations, the course of actions may change.

Novelty characterization and response generation sound extremely challenging. However, it is not impossible to do because in most applications the set of responses is finite. For example, in self-driving cars, the set of responses includes slowing down, stopping the car, and swerving. Based on the set of responses, we can work backward to build the needed systems to detect the characteristics of the detected novelty. See Section 6 for an example.

**Learning to respond.** In some situations, the system may not know how to respond to a novel object or situation. It may try any of the following ways.

(1) *Asking a human user*. In the case of the self-driving car, when it does not know what to do, it may ask the passenger using the interactive module *I* in natural language and then follow the instruction from the passenger and learn it for future use. For example, if the car sees a black patch on the road that it has never seen before, it can ask "*what is that black thing in front*?" The passenger may answer "*that is tar*." If there is no ready response, e.g., no prior information on tar, the system may progress with a further inquiry, asking the passenger "*what should I do*?"

(2) *Imitation learning*. On seeing a novel object, if the car in front drives through it with no issue, the car may choose the same course of action as well and also learn it for future use if the car drives through without any problem.

(3) *Reinforcement learning*. By interacting with the environment through trial-and-error exploration, the agent learns a good response policy. This is extremely challenging in a real-life environment as any action taken has consequences and cannot be reversed. For this to work, the agent must have an internal evaluation system that can assign rewards to states and assess risk or safety of each action.

(4) *Transfer learning*: The agent may transfer knowledge from previous similar environments to the new unknown environment. Note that researchers in the control community have worked on the topic of finding a feasible trajectory for a new task in an unknown environment. For example, Vallon and Borrelli (2020) proposed a hierarchical learning architecture for predictive control in unknown environments, which is based on generalization and knowledge transfer from previous familiar and similar environments. However, for such transfers to be successful, novelty characterization is critical; otherwise, knowledge transfer can be detrimental, e.g., resulting in negative transfer.

If multiple novelties are detected at the same time, it is more difficult to respond as the agent must reason over the characteristics of all novel objects to dynamically formulate an overall plan of actions that prioritizes the responses.

### 5.5. Risk Assessment and Learning

There is risk in achieving performance goals of an agent when making an incorrect decision. For example, classifying a known guest as unknown or an unknown guest as known may negatively affect guest impressions resulting in negative reviews. For a self-driving car, misidentifications can result in wrong responses, which could be a matter of life and death. Thus, risk assessment must be made in making each decision. Risk assessment can also be learned from experiences or mistakes. In the example of a car passing over tar, after the experience of passing over shiny black surfaces safely many times, if the car slips in one instance, the car agent must assess the risk of continuing the prior procedure. Given the danger, a car may weigh the risk excessively, slowing down on new encounters of shiny black surfaces.

Another aspect of risk is in the adaptation process. The planned or learned actions should be safe. In the control and reinforcement learning community, many safe reinforcement learning methods have been proposed. For example, Mazouchi et al. (2021) presented a conflict-aware safe reinforcement learning algorithm to control autonomous systems. Instead of providing safety and performance guarantees for a single environment or circumstance, this paper proposes a method to provide safety and performance guarantees across a variety of circumstances that the system might encounter.

## 6. CML: An Example SOLA System

Although novelty detection (Yang et al. 2021; Pang et al. 2021) and incremental or continual learning (Chen and Liu 2018; Parisi et al. 2019; Lomonaco et al. 2022) have been studied widely, little work has been done to build a SOLA system. Here we describe an implemented task-oriented dialogue system or chatbot, CML (*Command Matching and Learning*), that follows the SOLA framework. CML performs each function in SOLA continually by itself on the job during conversation. Below, we provide an overview of the system and highlight each corresponding component in the SOLA framework and discuss how it works. Details of the system and experimental evaluations can be found in (Mazumder et al. 2020b). Another two related systems that learn factual knowledge during conversation can be found in (Mazumder et al. 2019, 2020a).

CML[1] is a natural language interface (NLI) like Amazon Alexa and Apple Siri. Its ***performance task*** is to take a user command in natural language (NL) and perform the user requested API action in the underlying application. Since it is a text-based system, no other support function is needed. The key issue is how to understand paraphrased NL commands from the user to map a user command to a system's API call.

CML is based on *natural language to natural language* (NL2NL) matching to automatically build NLIs. The approach is application-independent and requires no pre-collected application-specific training data, and thus can be easily adapted to different applications, e.g., robot navigation and command systems, virtual assistants like Siri and Alexa, and GUI-based software applications (e.g., manipulating objects in MS Word, MS Paint, Windows). To build a new NLI (or to incrementally add a new task/skill to an existing NLI), the application developer only needs to write a set $S_i$ of seed commands (SCs) in NL to represent each API $a_i$ $\in A$ (which is the set of all API actions that can be performed in the application). SCs in $S_i$ are just like paraphrased NL commands from the end users to invoke $a_i$. The only difference is that the objects to be acted upon in each SC are replaced with variables, which are the arguments of API $a_i$. When the user issues a command *C*, the system simply matches *C* with a SC $s_k^*$ of the correct action $a^*$ and in doing so, it also instantiates the variables/arguments for the associated API $a^*$ to be executed. For example, the table below shows three API actions (column 1) for *switching on the light in a location*, *switching off the light in a location*, and *change the light color of a location*, respectively. Two example SCs are given for each API action in column 2. An example user command (without XI or X2) is given in column 3 for each API. Here, X1 and X2 are variables or place holders for *location* and *color* respectively.

[1] We note that many terminologies used here are different from those used the original CML paper. This is because when the CML paper was written, the SOLA framework had not been conceived yet. However, since CML is an open-world continual learning system, its steps and modules naturally map well to those of the SOLA framework.

| APR(arg: arg type) | Seed Commands (SCs) | Example User Command |
|---|---|---|
| SwitchOnLight(X1: location) | 1. Switch on the light in X1<br>2. Put on light in X1 | Switch on the light in the **bedroom (X1)**. |
| SwitchOffLight(X1: location) | 1. Switch off the light in X1<br>2. Put off light in X1 | Switch off the light in the **bedroom (X1)** |
| ChangeLightColor (X1: location, X2: color) | 1. Change the X1 light to X2<br>2. I want X1 light to be X2 | Change the **bedroom (X1)** light to **blue (X2)** |

CML has three main modules: (1) A *SC specification language* for the application developer to specify the initial SCs for its application, (2) a *command grounding module* (CGM) to match/ground a user command (e.g., "*power on the light in the bedroom*") to a SC (e.g., [Put on the light in X1], where the grounded API argument is {X1 = 'bedroom'}) for the associated API action (e.g., SwitchOffLight(X1:location)) to be performed, and (3) an *interactive learner* to continually learn new SCs from users during application.

***Novelty*** equates to the CGM's failure in grounding a user command. When the system detects a novelty (a hard-to-understand user command), it tries to understand the command and also learn the command so that it will be able to understand it and similar commands in the future. The system also assumes that every novelty is **relevant** to the application. The novelty ***characterization*** step of CML, which is also done by CGM, tries to identify the part of the user command that the system does not understand and how similar it is to some known commands. CGM uses an information retrieval (IR) based matching model. A Pre-trained language model can also be employed to build CGM.

Based on the characterization, the system ***adapts*** by asking the user via an interactive dialogue to obtain the ground truth API action requested by the user, which also serves as a piece of training data for ***continual learning***. In the adaptation or accommodation process, risk is also considered.

Consider the following example. The user issues the command "*turn off the light in the kitchen*" that the system does not understand (i.e., a **novelty**), i.e., the CGM module fails to ground/match the command. Based on the current system state, it decides which part of the command it can understand or ground, which part it has difficulty with, and what known commands are similar to the user command (i.e., **characterization**). Based on the characterization result, the interactive learner provides the user a list of top-*k* predicted actions (see below) described in NL and asks the user to select the most appropriate action from the given list (i.e., **adaptation**).

**User:** Turn off the light in the kitchen

**Bot:** Sorry, I didn't get you. Do you mean to:

**option-1**. switch off the light in the kitchen, or
**option-2**. switch on the light in the kitchen,
**option-3**. change the color of the light?

The user selects the desired action (option-1). The action API [SwitchOffLight(X1:location)] corresponding to the selected action (option-1) is retained as the ground truth action for the issued user command. In subsequent turns of the dialogue, the interactive learner will ask the user questions to acquire ground truth values associated with the arguments of the selected action, as defined in the API. This process is controlled by an action **planner**. CML then **incrementally learns** to map the original command "*turn off the light in* *the kitchen*" to the API action, SwitchOffLight(arg:location). Learning here means to create a new SC [turn off the light in X1] and add it to the list of SCs for the API SwitchOffLight(X1:location) so that in the future, when this user or any other user issues the same or a similar command, CML will have no problem in understanding or grounding it, i.e., mapping the command to this new SC. Over time, CML learns more and more from users and becomes more and more knowledgeable and powerful in serving them.

**Risk** is considered in CML in two ways. First, it does not ask the user too many questions in order not to annoy the user. Second, when the characterization is not confident, the system simply asks the user to say his/her command again in an alternative way (which may be easier for the system to ground or understand) rather than providing a list of random options for the user to choose from. If the options have nothing to do with the user command, the user may lose confidence in the system.

## 7 Key Challenges

Although novelty detection and continual learning have been researched extensively, they remain challenging. Limited work has been done to address the following (this list is by no means exhaustive):

**Learning everything and everywhere.** As indicated earlier, every module or sub-system in Figure 2 needs to learn continually, i.e., everywhere needs learning. Similarly, everything can be learned. For example, from each user's dialogue history with a chatbot, the system can learn whether a user feels more excited or gets annoyed while conversing on a particular topic, and what he/she likes and dislikes. The chatbot can then utilize this user's profile in modeling future conversations to make them more engaging with the user.

In this paper, we focus only on the open world continual learning of the main task. A general framework is needed to integrate all the learning activities and their resulting knowledge to make the agent work even better.

**Obtaining training data on the fly.** A key feature of SOLA is the interaction with human users to obtain ground-truth training data, which needs a dialogue system. Building an effective dialogue system for this purpose is challenging. We are unaware of any such system for SOLA except CML (Mazumder et al. 2020b), but CML is only for simple command learning.

**Few-shot continual learning.** It is unlikely for the learning agent to collect a large volume of training data via interaction with the user. Then, an effective and accurate few-shot continual/incremental learning method is necessary.

**Novelty characterization and adaptation.** Characterization is critical because it defines the characteristics used to recognize world state and determine the best response strategy. We have given several examples of characterization and adaptation in the domains of self-driving cars and greeting bots. However, little research has been done on the topics in the academic community. They are extremely challenging as they require the system to have a large amount of prior knowledge and a domain world model, and to reason based on this knowledge and the current observations.

**Learning to respond/adapt.** As indicated earlier, this is especially challenging in a physical environment (Dulac-Arnold et al. 2021). For example, due to safety concerns, learning during driving by a self-driving car using reinforcement learning (RL) is very dangerous because every action has potentially life and death consequences and cannot be undone. Furthermore, for RL to work, the agent must have a highly effective internal reward or evaluation system to assign rewards to actions and states and be aware of safety constraints automatically without detailed manual specifications. So far limited work has been done.

**Knowledge representation, reasoning, and revision.** SOLA has so many components, but it is not known what knowledge representation & reasoning scheme best suits all modules and facilitates their integration. Further, it is inevitable that the system may misinterpret, generalize or otherwise assemble incorrect knowledge. A system must have a mechanism to detect and revise the inaccurate knowledge on its own. Little work has been done in this area.

## 8. Conclusion

A truly intelligent system must be able to learn autonomously and continually in the open world on its own initiative after deployment, adapt to the ever-changing world, and learn more and more to become more and more powerful over time. This paper proposed the self-initiated open-world continual learning and adaptation (SOLA) framework for this purpose, and presented the concepts, steps, a general framework and key challenges. An implemented SOLA system called CML in the context of dialogue systems (or chatbots) was also described. We believe that future research in SOLA will bring AI to the next level.

## Acknowledgments

This paper benefited from numerous discussions in DARPA SAIL-ON Program meetings and. This work was supported in part by a DARPA Contract HR001120C0023. Bing Liu was also partially supported by three National Science Foundation (NSF) grants (IIS-1910424, IIS-1838770 and CNS-2225427), and a research gift from Northrop Grumman. *The views expressed in this document are entirely those of the authors and not those of the funders or the authors' respective organizations.*

## Authors' Biography

**Bing Liu** is a distinguished professor of Computer Science at the University of Illinois at Chicago (UIC). He received his Ph.D. in Artificial Intelligence (AI) from the University of Edinburgh. His research interests include lifelong and continual learning, sentiment analysis, lifelong learning chatbots, open-world AI/learning, NLP, and machine learning. He has published extensively in top conferences and journals and authored four books: one about lifelong learning, two about sentiment analysis and one about Web mining. Three of his papers received Test-of-Time awards: two from SIGKDD and one from WSDM. Another of his papers received Test-of-Time award - honorable mention from WSDM. Some of his work has been widely reported in the international press, including a front-page article in The New York Times. He has served as the Chair of ACM SIGKDD from 2013-2017, as program chair of many leading data mining conferences, including KDD, ICDM, CIKM, WSDM, SDM, and PAKDD, as associate editor of leading journals such as TKDE, TWEB, DMKD and TKDD, and as area chair or senior PC member of numerous NLP, AI, Web, and data mining conferences. He is the winner of 2018 ACM SIGKDD Innovation Award. He is a Fellow of the ACM, AAAI, and IEEE.

**Sahisnu Mazumder** is an AI Research Scientist at Intel Labs, USA where he works on Human-AI collaboration and dialogue & interaction systems research. Prior to joining Intel, he obtained his PhD in Computer Science at the University of Illinois at Chicago (UIC), USA. His research interests broadly falls into the area of NLP, Deep Learning, Dialogue and Interactive Systems, Lifelong and Continual Learning, Open-World AI / Learning. He has published several research papers in leading AI, NLP and Dialogue conferences like AAAI, IJCAI, ACL, EMNLP, NAACL, SIGDIAL; delivered tutorials in SIGIR-2022, IJCAI-2021, BDA-2014; served as Reviewer of premier conferences like AAAI, IJCAI, ACL, EMNLP, NAACL, EACL, and journals like ACM TALLIP, IEEE TNNLS. He also worked as a Research Intern at Huawei Research USA on projects related to user activity & interest mining and at Microsoft Research - Redmond on Natural Language Interaction (NLI) system design.

**Eric Robertson** received his Masters in Computer Science from UMBC, and has over 30 years of experience in complex event processing, data science, geospatial analytics, ontologies, and large-scale data systems across multiple industries including Financial, Telecommunication, Pharmaceutical and Defense. He served as PAR Government's PI on the DARPA AIDA and SAIL-ON programs and has led development of the data collection and evaluation infrastructure for the DARPA MediFor (Media Forensics) program. He holds three patents in Cyber Intrusion Detection Systems.

**Scott Grigsby** received his Bachelor's degree in Physics from Lycoming College and his PhD from The Ohio State University in Biophysics. Dr. Grigsby has over 30 years' experience in Cognitive Science including extensive experience in science and engineering related to autonomous systems, artificial and augmented intelligence, human-machine teaming, human interfaces, simulation and training, decision support, and command and control technologies and has published over 50 papers in related areas.